\documentclass{article}

\PassOptionsToPackage{numbers, sort&compress}{natbib}

\usepackage{graphicx}

\usepackage[main, final]{neurips_2026}

\usepackage[utf8]{inputenc} 
\usepackage[T1]{fontenc}    
\usepackage{hyperref}       
\usepackage{url}            
\usepackage{booktabs}       
\usepackage{amsfonts}       
\usepackage{nicefrac}       
\usepackage{microtype}      
\usepackage{xcolor}         
\usepackage{tikz}
\usetikzlibrary{arrows.meta, positioning}
\usepackage{siunitx}
\usepackage{booktabs}
\usepackage{multirow}
\usepackage{booktabs}
\usepackage{array}
\usepackage{makecell}
\usepackage{amsmath, amssymb, bm}
\usepackage{booktabs}

\usepackage{eso-pic}
\definecolor{ourcolor}{RGB}{0,128,0} 

\title{Streaming-AAA: \\ Streaming AAA Game Video Generation at \\ High Resolution, High Frame Rate, and Low Latency}

\author{
  Zhihao Hu*$^\dagger$,
  Jianqi Ma*,
  Jun Zhang$^\dagger$,
  Jiayang Bai,
  \\
  \textbf{Haifeng Shi, Yanxiang Zhuang, Zhongqian Sun, Weidong Zhang, Wei Yang} \\
  Tencent AIPD \\
  Project Page: \url{https://streamingaaa.github.io/}
}

\begin{document}
\maketitle

\begin{tikzpicture}[remember picture, overlay]
    \node[anchor=north east, xshift=-3.8cm, yshift=-1.6cm] 
        at (current page.north east) {
            \includegraphics[height=1.0cm]{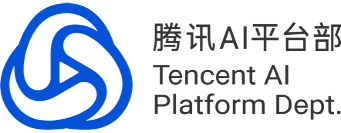}
        };
\end{tikzpicture}

\vspace{-7mm}
\begin{figure}[!h]
    \centering
    \includegraphics[width=0.95\textwidth]{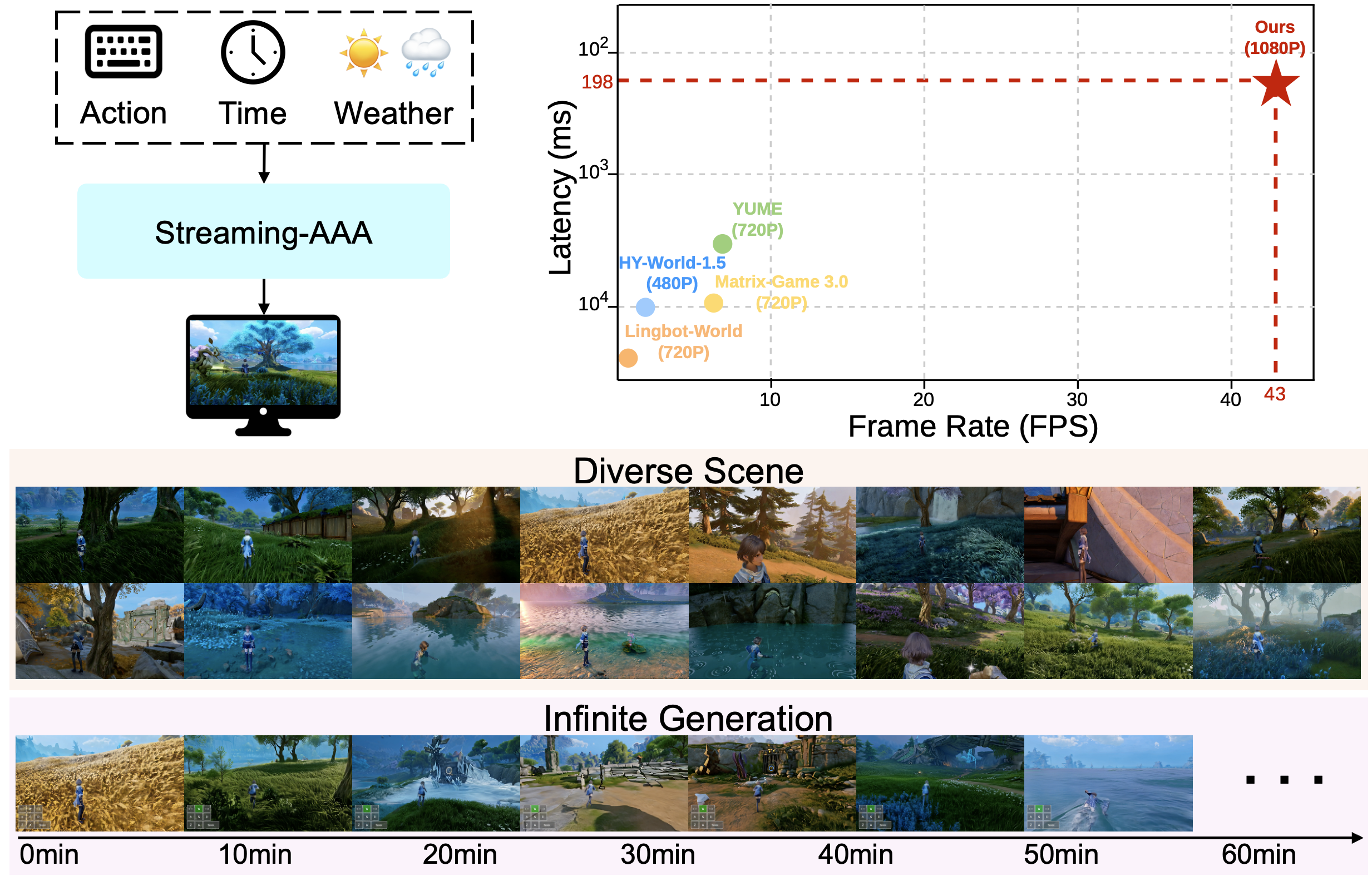}
    \caption{Overview of Streaming-AAA: Streaming AAA game video generation framework applied to \textit{Genshin Impact}, demonstrating high-resolution, high-frame-rate, and low-latency gameplay synthesis. Speed and latency are tested on NVIDIA A800 GPUs.}
    \label{fig:teaser}
\end{figure}

\renewcommand{\thefootnote}{}
\footnotetext{* Equal contribution. $^\dagger$ Project Lead.}
\renewcommand{\thefootnote}{\arabic{footnote}}

\begin{abstract}
Controllable streaming video generation offers a promising route toward interactive game simulation, yet existing methods remain far from practical gameplay. They often fail to jointly achieve high visual fidelity, high frame rates, and low interaction latency, resulting in poor responsiveness and degraded user experience. We present Streaming-AAA, a new paradigm for controllable infinite generation of AAA-quality game videos. Our system generates \textbf{1080p} gameplay videos at \textbf{43 FPS} with \textbf{198 ms} end-to-end latency interaction on merely \textbf{NVIDIA A800 GPUs}, while supporting long-horizon play without severe temporal degradation or collapse.
Streaming-AAA is built upon three key designs. First, we redesign the VAE architecture for efficient high-resolution decoding, enabling real-time synthesis at 1080p. Second, we pretrain the diffusion model using diffusion forcing and significantly extend the temporal training horizon, allowing the model to capture long-term dependencies and maintain persistent memory. Third, we perform post-training distillation and memory compression to improve inference efficiency while retaining visual quality and temporal consistency. Experiments show that Streaming-AAA substantially improves visual fidelity, frame rate, latency, and long-term stability over prior controllable video generation approaches, enabling a significantly more playable experience for interactive AAA game video generation. 
\end{abstract}

\section{Introduction}
\label{sec:intro}

Interactive game simulation has become an increasingly important application of generative video models, where the system must synthesize future visual observations conditioned on user controls in a streaming manner~\cite{ha2018world,ali2025world,sun2025worldplay,hong2025relic,mao2026yume1}. Unlike offline video generation, playable game generation imposes simultaneous requirements on visual fidelity, temporal consistency, frame rate, and interaction latency. These requirements are especially stringent for AAA-level games, where high-resolution rendering, fast motion, complex scenes, and long-horizon interactions are all essential to the user experience. In this work, we primarily target \textit{Genshin Impact}, a widely played open-world action RPG whose environments are familiar enough that generation quality is easy to judge.

Despite recent progress in controllable video generation and world models~\cite{yu2025gamefactory,genie3,mao2026yume1,he2025matrix,tang2025hunyuan,ye2025yan,hyworld2025}, existing streaming generation systems remain far from practical gameplay. Many approaches can produce visually plausible short clips, but their quality degrades over long horizons. Others support control or interaction but operate at low resolution, low frame rate, or high latency, making the generated experience difficult to play. As a result, current methods do not jointly satisfy the key requirements of high-fidelity 1080p generation, real-time frame rates, low end-to-end latency, and stable infinite interaction.

We present \textbf{Streaming-AAA}, a controllable infinite streaming AAA game video generation framework designed for high-resolution, high-frame-rate, and low-latency gameplay. Streaming-AAA first introduces an efficient high-resolution video VAE that compresses 1080p gameplay frames into compact spatio-temporal latents and reconstructs high-fidelity frames with low decoding overhead. On top of this representation, we build a UNet-based latent diffusion model with factorized spatial, temporal, and cross-attention modules, enabling efficient visual modeling, temporal reasoning, and multi-modal control. To support long-horizon streaming generation, we further adopt long-context pre-training, post-training alignment with distribution matching distillation, and a hybrid short-term and long-term KV-cache memory. These designs allow Streaming-AAA to preserve long-range consistency while maintaining low interaction latency during real-time inference.

Our main contributions are summarized as follows:
\begin{itemize}

    \item We propose \textbf{Streaming-AAA}, the first controllable infinite-horizon AAA game video generation framework that simultaneously delivers high visual fidelity, real-time throughput (\textbf{\SI{43}{FPS}} at \textbf{1080p}), ultra-low end-to-end latency (\textbf{\SI{198}{ms}}), and stable long-rollout consistency. To our knowledge, this is the first system to achieve production-grade real-time gameplay generation at this specification.
    
    \item At its core, Streaming-AAA comprises a purpose-built causal video VAE and a factorized 3D-UNet latent diffusion backbone. The VAE compresses raw pixel inputs by 16~$\times$ spatially into compact latent representations, drastically reducing inference overhead. The 3D-UNet decouples spatial texture modeling, temporal dynamics learning, and conditional control via cross attention modules cutting computational cost, while supporting flexible integration of rich frame-aligned conditioning signals (e.g., agent actions, weather states).
    
    \item We establish a two-stage pre/post-training paradigm tailored for unbounded autoregressive streaming. In pre-training, we adopt a per-frame independent noise scheduling strategy (i.e., Diffusion Forcing) under the DDPM framework, which encodes the inductive bias of growing uncertainty for distant temporal horizons to match open-world evolution dynamics. In post-training, we apply distribution matching distillation~(DMD) to align the model with \textbf{2-step} generation, which further boosts rollout stability and temporal consistency without sacrificing real-time performance.

\end{itemize}
\section{Related Works}
\label{sec:related}

\subsection{Video Generation Models}
Recent years have witnessed remarkable progress in video generation. Large-scale text-to-video diffusion models~\cite{gupta2024photorealistic,gao2025seedance,seedance2026seedance,kong2024hunyuanvideo,team2025longcat,wan2025wan,zhang2025waver,contributors2025lightx2v,hacohen2026ltx,Sora,team2025kling,teng2025magi} have demonstrated impressive visual quality and diversity, producing photorealistic short clips from textual descriptions. However, these models typically require minutes of inference time to generate only a few seconds of video, and most of them lack the ability to produce arbitrarily long sequences in a streaming fashion.

To address the long-horizon limitation, several methods\cite{huang2026self,cui2025self,chen2026context,yang2025longlive,chen2024diffusion,cui2026lol,zhu2026causal} attempt to retrofit pre-trained text-to-video models for streaming generation. Self-Forcing~\cite{huang2026self} bridges the train-test gap in autoregressive video diffusion by aligning the training distribution with the streaming rollout. Context Forcing~\cite{chen2026context} introduces long-context conditioning to maintain consistency across extended autoregressive generation. LongLive~\cite{yang2025longlive} targets real-time interactive long video generation by combining efficient architectures with streaming inference. Alternatively, Diffusion Forcing~\cite{chen2024diffusion} equips the model with streaming capability directly during pre-training by combining next-token prediction with full-sequence diffusion. Despite these advances, achieving real-time streaming inference at high resolution remains a significant challenge for playable game generation.

\subsection{World Models}
World models aim to learn predictive representations of environments for planning and interaction. Existing approaches fall into two categories.

The first is based on 3D reconstruction~\cite{yu2024wonderjourney,yu2025wonderworld,Marble}, which builds explicit representations such as 3D Gaussian Splatting for real-time exploration. While achieving strong real-time performance, these methods produce mostly static worlds with limited interactivity.

The second leverages video generation for world simulation. Early works~\cite{bar2024lumiere,blattmann2023stable,brooks2024video,decart2024} demonstrate that video diffusion models can serve as world simulators. More recently, a growing body of research~\cite{ha2018world,assran2025v,bruce2024genie,hong2025relic,ali2025world,sun2025worldplay,xiang2025pan,wang2026worldcompass,chen2025deepverse,gao2026dreamdojo,hafner2025training,yang2024playable,savva2026solaris} has explored interactive and controllable world models, addressing challenges such as long-horizon memory, action conditioning, and real-time streaming. However, none of these methods simultaneously achieve high resolution (1080p), high frame rate, and low interaction latency—all essential for AAA-level playable game generation. Our work addresses this gap by jointly optimizing visual fidelity, streaming efficiency, and interactive responsiveness within a unified framework.

\section{Dataset Construction}
\label{sec:dataset}
Modern AAA video games represent a convergence of cutting-edge graphics technologies, providing an almost inexhaustible source of high-fidelity visual data. Specifically, these environments feature:
\begin{itemize}
    \item \textbf{Geometrically Diverse Landscapes:} Expansive terrains encompassing natural formations (mountains, rivers, oceans, forests, grassland) and intricate man-made architectures.
    \item \textbf{Complex Human-Scene Interactions:} Rich agent behaviors such as locomotion across uneven terrain, climbing trees, and swimming in dynamic fluid simulations.
    \item \textbf{Sophisticated Lighting Conditions:} Dramatic variations in illumination, including dawn, noon, and night cycles, often coupled with advanced effects like ray-traced global illumination.
    \item \textbf{Realistic Physical Dynamics:} Natural phenomena such as wind-induced vegetation swaying (wind-sway effects) and physically-based object interactions.
\end{itemize}
To train a generative model capable of synthesizing photorealistic videos with such intricate spatio-temporal details, it is \textit{imperative} to expose the model to a massive and highly diverse dataset that encapsulates these multifaceted distributions. However, manual gameplay recording is labor-intensive and suffers from limited scalability, making it infeasible to cover the vast state space of modern AAA titles.

To bridge this gap, we design an automated parallel data acquisition system, termed Automated Data Collection System to efficiently harvest these rich visual cues. 

\subsection{Automated Data Collection System (ADCS)}

\begin{figure}[t]
    \centering
    \includegraphics[width=\linewidth]{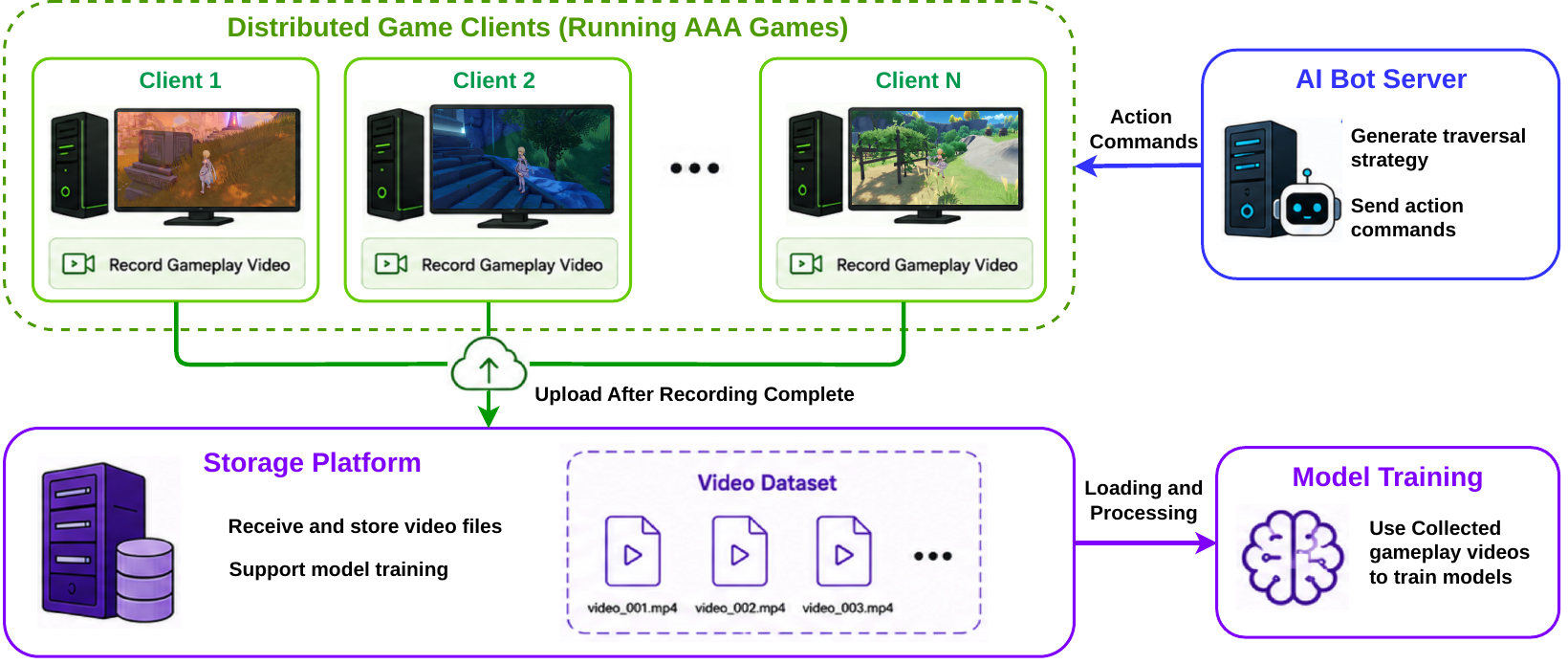}
    \caption{\textbf{Overview of the ADCS.} Designed to capture rich features of AAA games—including diverse landscapes, dynamic interactions (e.g., climbing, swimming), and photorealistic lighting—the system achieves a daily throughput of >10M frames and densely covers KM$^2$s of virtual terrain within several days.}
    \label{fig:pipeline}
\end{figure}

As illustrated in Figure~\ref{fig:pipeline}, the proposed Automated Data Collection System (ADCS) comprises three integral components: an \textbf{AI Agent Server}, a \textbf{Distributed Render Cluster}, and a \textbf{Centralized Storage Backend}.

\noindent\textbf{AI Agent Server and Control Logic.}
The AI Agent Server acts as the control center, orchestrating in-game agents in Distributed Render Cluster via the LAN. It dispatches low-level actions to each PC game instance. To ensure comprehensive environmental coverage, we implement an autonomous navigation policy that samples actions to sufficiently cover the virtual terrains.

\noindent\textbf{Distributed Render Cluster.}
The cluster consists of a pool of high-end workstations executing game instances in parallel. Each node captures raw gameplay at $30$ FPS and 1080p resolution using OBS Studio, preserving fine-grained motion details essential for learning physical dynamics and temporal coherence. Frame-aligned agent actions and real-time frame rate are recorded alongside the video stream. Weather and time-of-day are not read from the game; instead, they are estimated by a model~(\textit{i.e.}, open-source Qwen-VL~\cite{bai2023qwen}) from the captured frames. Each recording task lasts $60$ seconds. And when each task ends, the recorded video along with its annotations will be upload to the Centralized Storage Backend.

\noindent\textbf{Centralized Storage Backend.}
The backend serves as the unified data hub, ingesting asynchronously uploaded packages from all render nodes upon completion of each capture session. Each upload contains synchronized pairs of high-fidelity gameplay videos and their corresponding metadata streams extracted by the render cluster. The curated dataset is directly accessible to the training cluster for distributed training.

\noindent\textbf{Scalability and Throughput.}
The tight integration of the three components underpins the pipeline’s exceptional efficiency. By leveraging parallel execution across the render cluster and high-throughput asynchronous data ingestion at the storage backend, the system collects over \num{5} million frames daily. Within days of acquisition cycle, we successfully captured footage spanning approximately $5$ KM$^2$ of in-game terrain. This automated infrastructure ensures a steady supply of high-quality, curated data, enabling the model to effectively learn the complex distributions inherent in AAA game environments.

\subsection{Conquering Huge Map with ADCS}

Modern AAA titles are characterized by their sheer geographical scale, often featuring seamless open worlds spanning square kilometers~(KM$^2$). 
Naïvely employing a sequential, single-agent collection strategy (\textit{i.e.}, traversing the map linearly) presents a significant bottleneck: the data acquisition rate is fundamentally constrained by the agent's movement speed and loading times. This results in \textbf{prohibitively low efficiency} and insufficient \textbf{spatial data density}, failing to meet the stringent demands of training high-fidelity video world models which require dense observations of environmental geometry and dynamics.

To address this challenge, we leverage our proposed ADCS to achieve high-density spatial coverage. 
Our core strategy involves decoupling spatial navigation from data capture through parallelization and teleportation.

\noindent\textbf{Parallel Spatial Sampling Strategy.}
To ensure comprehensive coverage and high sampling density across expansive virtual environments, we employ a parallel spatial sampling strategy. 
First, we partition the game map with the \textit{teleportation waypoints} in the map. 
 
Rather than relying on conventional pathfinding to traverse the terrain sequentially—which is time-prohibitive for large-scale maps—we leverage the game's built-in \textit{teleportation waypoint}. 
Specifically, the AI Agent Server dispatches specific \textit{teleportation waypoint} targets to the Distributed Render Cluster, routing the in-game avatars to the designated waypoints.

This approach allows us to distribute data collection tasks across multiple compute nodes in parallel. Each client is assigned a subset of the waypoints, recording high-frame-rate footage locally upon arrival. 
By decoupling spatial navigation from data capture, our strategy significantly accelerates the acquisition process and enhances the spatial density of the dataset. Consequently, the system can efficiently survey approximately $5$ KM$^2$ of terrain in several days, capturing fine-grained environmental details without the computational overhead of simulating long-distance traversal.

\subsection{Data Statistics}
Leveraging our proposed \textbf{ADCS} and the parallel spatial sampling strategy, we construct a large-scale, high-fidelity dataset that captures the rich visual and interactive complexity of modern AAA games. 
In total, we collect approximately $22,000$ videos, each lasting $1$ minute and recorded at $30$ fps, aggregating to approximately $40$ million frames. All footage is recorded at a resolution of $1920 \times 1080$, ensuring fine-grained spatial and temporal details suitable for high-capacity video generation models.

Beyond raw pixel data, our collection pipeline provides frame-aligned annotations that are typically unavailable in real-world datasets. 
Specifically, each frame is annotated with:
\begin{itemize}
    \item \textbf{Agent Actions:} Control signals~("W", "A", "S", "D" to move, $\uparrow$ $\downarrow$ $\leftarrow$ $\rightarrow$ to turn, "Space" to jump), recorded directly from the game agent.
    \item \textbf{Environmental Context:} Weather conditions and diurnal lighting cycles (time-of-day), which are \textit{estimated by a vision-language model} (Qwen-VL~\cite{bai2023qwen}) rather than read from the game engine.
\end{itemize}

The data acquired through our ADCS framework exhibits richness across three distinct dimensions, each contributing to the overall fidelity and utility for training high-capacity video generation models.

\noindent\textbf{1. High-Fidelity Visual Stream.}
All videos are recorded at a resolution of $1920 \times 1080$ with a frame rate of $30$ fps. This high-resolution, high-frame-rate capture ensures access to fine-grained visual details and temporally coherent motion dynamics—essential for modeling complex scene evolution and object interactions.

\noindent\textbf{2. Rich Metadata Annotations.}
Each frame is accompanied by annotations, including recorded control signals and model-estimated environmental context (weather and time-of-day). These annotations enable fine-grained controllability during training and inference, allowing the model to condition its output on specific physical or semantic states.

\noindent\textbf{3. Broad Coverage with High Density.}
Through parallelized spatial sampling, we achieve extensive geographic coverage ($5$ KM$^2$) while maintaining high sample density. This combination ensures exposure to a diverse array of content including natural scenery and man-made architectures, enhancing the generalizability of learned representations.

These annotations facilitate a wide range of downstream tasks, from conditional video synthesis to world model training that requires strict physical and geometric consistency. Statistics of the dataset are summarized in Table~\ref{tab:dataset_stats}.

\begin{table}[t]
    \centering
    \caption{\textbf{Statistics of the Collected AAA Game Dataset.} Our dataset provides high-resolution video paired with rich metadata.}
    \label{tab:dataset_stats}
    \begin{tabular}{@{}lr@{}}
        \toprule
        \multicolumn{2}{c}{\textbf{Dataset Overview}} \\ 
        \midrule
        Total Videos & $\sim$\num{22}K \\
        Video Duration & 1 min \\
        Total Frames & $\sim$\num{40}M \\
        Resolution & $1920\times1080$ \\
        Frame Rate & 30 \\
        Coverage Area & > 5 KM$^2$ \\
        \midrule
        \multicolumn{2}{c}{\textbf{Annotations}} \\
        \midrule
        \multicolumn{2}{l}{\quad $\bullet$ Agent Actions (Movement, Interaction)} \\
        \multicolumn{2}{l}{\quad $\bullet$ Weather, Time-of-Day (Model-Estimated)} \\
        \bottomrule
    \end{tabular}
\end{table}

\section{Method}
\label{sec:method}

\subsection{Problem Formulation}
We formulate controllable AAA game video generation as an infinite streaming prediction problem. Given previous frames $x_{1:t}$, user controls $a_{1:t+k}$, and a compact memory state $m_t$, the model predicts the next video segment:
\begin{equation}
    p_{\theta}(x_{t+1:t+k} \mid x_{1:t}, a_{1:t+k}, m_t).
\end{equation}
Unlike offline video generation, this task requires causal generation, faithful response to user inputs, and long-horizon stability under strict constraints on resolution, frame rate, and interaction latency.

\subsection{Overview of Streaming-AAA}
We propose \textbf{Streaming-AAA}, a controllable infinite AAA game video generation framework for real-time gameplay. Streaming-AAA consists of three key components. First, an efficient high-resolution video VAE compresses 1080p frames into compact spatio-temporal latents and decodes them with low overhead. Second, a UNet-based latent diffusion model generates future video latents conditioned on player controls and historical memory. Third, long-context pre-training, post-training alignment, and hybrid KV-cache memory improve long-term consistency and reduce inference latency. Together, these designs enable high-fidelity, controllable, and low-latency streaming generation.

\subsection{Efficient High-Resolution Video VAE}
High-resolution decoding is a major bottleneck for real-time game video generation. We therefore redesign the VAE architecture to improve decoding efficiency while preserving fine visual details. The encoder compresses 1080p gameplay frames into compact spatiotemporal latents, and the decoder reconstructs high-fidelity frames with reduced computational overhead. This design allows the diffusion model to operate in latent space while enabling fast reconstruction of AAA-quality game visuals during streaming inference.

To bridge the gap between high-resolution spatial modeling and temporal consistency, we propose an Enhanced VAE model design. Our architecture is designed to efficiently handle ultra-high-definition inputs ($1792 \times 1024$) while maintaining a manageable latent space dimension with high-frequency details.

\noindent\textbf{Architecture Design.}
We configure an overall temporal-spatial compression factor of $4 \times 16 \times 16$ relative to the input resolution. The encoder maps the input video clip $x \in \mathbb{R}^{(4N+1) \times H \times W \times 3}$ to a latent representation $z \in \mathbb{R}^{(N+1) \times \frac{H}{16} \times \frac{W}{16} \times C}$, which is compatible for both single frame and multi-frame compression and decoding, similar to Wan-series. A key distinction of our model lies in the decoder's upsampling pipeline. Unlike traditional methods relying solely on interpolation or transposed convolutions, we introduce an additional \textbf{Pixel Shuffle} layer at the tail of the decoding process. Specifically, the decoder consists of three sequential Conv3D upsampling blocks followed by a \textbf{Pixel Shuffle} operation for efficient sub-pixel upsampling in the largest upsample stage. This design mitigates checkerboard artifacts commonly induced by large-stride deconvolutions and preserves fine-grained texture details.


\begin{figure}[tb]
  \centering
  \includegraphics[width=\linewidth]{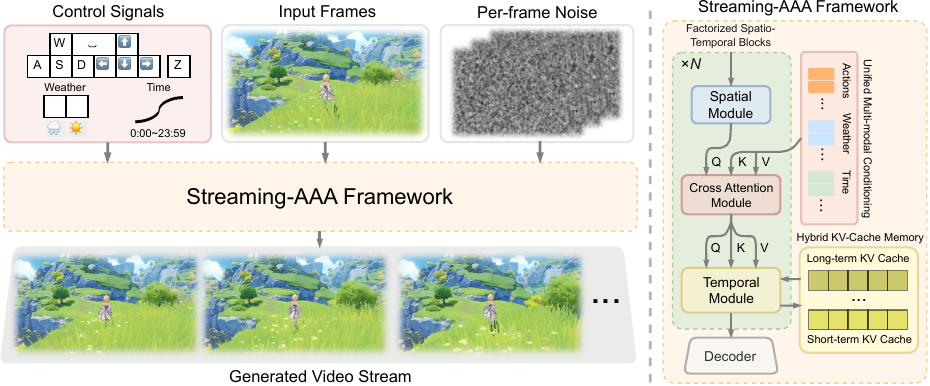}
  \caption{\textbf{Overview of the inference pipeline.} During inference, the 3D UNet-based Diffusion Model takes noise as input, injects control signals via cross-attention, and propagates historical frame information through temporal attention. After multiple denoising steps, the reconstructed latents are decoded by the VAE decoder to produce the generated video frames.}
    \vspace{-1mm}
  \label{fig:framework}
\end{figure}

\noindent\textbf{Multi-Stage Curriculum Training.}
Training a VAE from scratch for high-resolution video is inherently unstable. To address this, we employ a three-stage curriculum learning strategy for faster convergence:
\begin{itemize}
    \item \textbf{Stage 1 (Spatial Initialization):} We train the model on static images at $256 \times 256$ resolution to stabilize the reconstruction of basic color distributions and edges.
    \item \textbf{Stage 2 (Temporal Extension):} We extend the model to video data using $256 \times 256$ patches. Here, the model processes multiple frames within a clip to learn short-term temporal coherence.
    \item \textbf{Stage 3 (Native Resolution):} Finally, we fine-tune the model on full-resolution $1792 \times 1024$ videos to adapt to the aspect ratio and scale of real-world high-definition content.
\end{itemize}

\noindent\textbf{Rolling Frame Prediction.}
For multi-frame training, we adopt a rolling prediction mechanism. Given a sequence of length $T$, the model predicts frames indexed as $4n+1$ ($n \ge 0$). To balance computational cost and gradient flow, we selectively apply gradients only to the last 1 or 4 predicted frames of each sequence, preventing error accumulation in long sequences while maintaining temporal smoothness.

\subsection{Redesigning UNet-based Video Diffusion Models}

Leveraging the compact latent space provided by our VAE, we construct a 3D UNet-based diffusion model for high-fidelity video synthesis. 
The network follows a symmetric encoder-decoder structure, performing three downsampling and three upsampling operations, with the bottleneck resolution reduced to $\frac{1}{8}$ of the input.

\noindent\textbf{Factorized Spatio-Temporal Blocks.}
To mitigate the computational overhead of dense 3D operations, each stage of the UNet is composed of three specialized modules:
\begin{itemize}
    \item \textbf{Spatial Module:} Consisting primarily of convolutional layers, this module focuses on modeling intra-frame textures and structural details, ensuring high visual fidelity.

    \item \textbf{Cross-Attention Module:} Also constructed with Transformer layers, this module serves as the conditioning interface. It projects external signals into the latent space via cross-attention mechanisms in causal manner. This design allows the model to flexibly incorporate diverse control signals, such as \textit{action}, \textit{weather} and \textit{time-of-day} conditions, enabling fine-grained controllability over the generation process.

    \item \textbf{Temporal Module with Hybrid KV-Cache Memory:} Built upon Transformer blocks, this module operates along the temporal axis to capture the dynamic evolution of each latent pixel across frames in causal manner, ensuring motion consistency. We also implement a Hybrid KV-Cache Memory mechanism. 
    Once a frame is finalized, its corresponding keys and values in the self-attention layers are stored in memory. For subsequent frames, the model retrieves these cached key-value~(KV) pairs as tokens reference instead of recomputing them, significantly reducing the computational complexity from $O(T^2)$ to $O(T)$ per step, where $T$ is the sequence length. We select both specific long-term and short-term KV-tokens to enhance long-term temporal consistency within only \num{30} tokens.
    
\end{itemize}
This factorized architecture effectively decouples spatial modeling, temporal dynamics, and conditional control, significantly reducing the complexity compared to full spatio-temporal attention while enhancing generalization across various scenarios.

\noindent\textbf{Unified Multi-Modal Conditioning.}
To handle heterogeneous inputs, we adopt a unified conditioning scheme. 
Each condition type (e.g., action, weather, time, frame-rate) is first processed by a dedicated embedding layer and tokenized into a fixed-dimensional vector. 
Subsequently, these tokens are concatenated into a unified condition sequence. 
This sequence is then injected into the model through the Cross-Attention modules at each stage. 
This design not only standardizes the input interface for various modalities but also allows the model to selectively attend to relevant conditions during different stages of denoising.

\noindent\textbf{Training and Convergence.}
By operating within the compact latent space provided by our VAE, the diffusion model achieves stable convergence. The efficient design ensures a high inference throughput, making it suitable for practical applications requiring multi-conditional control.

\subsection{Modeling Long-horizon Uncertainty with Diffusion-forcing Pre-training}

We adopt the Diffusion Forcing (DF)~\cite{chen2024diffusion} paradigm in the pretraining stage of our streaming-AAA model pretraining, whose core innovation of per-frame independent noise scheduling aligns naturally with the requirements of long-horizon, controllable world simulation. Instead of enforcing a globally shared noise level across all frames, DF samples an independent noise step $k_i \sim p(k)$ (e.g., uniform or logit-normal distribution over the DDPM~\cite{ho2020denoising} timestep range $[0, K]$) for each frame $i$, where the per-frame noised latent follows the standard DDPM forward formulation:
\begin{equation}
    \mathbf{z}_i^{k_i} = \sqrt{\bar{\alpha}_{k_i}} \cdot \mathbf{z}_i^0 + \sqrt{1-\bar{\alpha}_{k_i}} \cdot \boldsymbol{\epsilon}_i, \quad \boldsymbol{\epsilon}_i \sim \mathcal{N}(\mathbf{0}, \mathbf{I}),
\end{equation}
with $\mathbf{z}_i^0$ denoting the clean VAE-encoded frame latent, and $\bar{\alpha}_{k_i} = \prod_{s=1}^{k_i} (1-\beta_s)$ representing the cumulative product of noise retention coefficients under a predefined DDPM schedule (e.g., cosine schedule), where $\beta_s \in (0,1)$ is the pre-defined noise addition ratio at step $s$. Stacking these per-frame noised latents forms a 2D noise schedule matrix that encodes the inductive bias of growing uncertainty for farther temporal horizons—matching the intrinsic causal dynamics of open-world physical evolution. This design, combined with \textit{velocity (v-)prediction}, enables stable arbitrary-length rollouts. Crucially, the per-frame noise assignment is natively compatible with the rich, frame-aligned metadata, enabling frame-level acute dynamic variation and transition modeling.

For diffusion training, we opt for v-prediction instead of conventional $\bm{\epsilon}$-prediction. The velocity $v_i$ for frame $i$ is defined as the linear combination of the clean latent and noise term in the forward process:
\begin{equation}
    v_i = \sqrt{\bar{\alpha}_{k_i}} \cdot \boldsymbol{\epsilon}_i - \sqrt{1-\bar{\alpha}_{k_i}} \cdot \mathbf{z}_i^0,
\end{equation}

Let $\mathbf{z}^{\mathbf{k}} \triangleq \{\mathbf{z}_i^{k_i}\}_{i=1}^{T}$ be the whole noised latent sequence, $\mathbf{k}=\{k_i\}_{i=1}^{T}$ the per-frame noise steps, and $\mathbf{c}$ the control condition sequence (aggregating action, weather, time, and frame-rate signals). We denote $f(\cdot)_i$ as the $i$-th frame of its output, and optimize the per-frame v-prediction objective:
\begin{equation}
\label{eq:df}
    \mathcal{L}_{\text{DF}} = \mathbb{E}_{\{\mathbf{z}_i^0\}, \mathbf{c}, \mathbf{k}, \{v_i\}} \;
    \frac{1}{T} \sum_{i=1}^T
    \left\| f\!\left(\mathbf{z}^{\mathbf{k}}, \mathbf{c}, \mathbf{k}\right)_i - v_i \right\|_2^2,
\end{equation}
where $T$ is the sequence length. This contrasts with the unified-timestep v-prediction loss that enforces a single noise level $t$ for all frames. The per-frame independent $(\mathbf{z}_i^{k_i}, k_i, v_i)$ pairs act as an implicit curriculum, stabilizing pretraining on our heterogeneous AAA dataset and enabling continuous post-training distributions for infinite rollouts. Empirically, v-prediction mitigates error accumulation during long-horizon generation compared to $\bm{\epsilon}$-prediction, further reinforcing the stability of our streaming inference pipeline.

\noindent\textbf{Stage 1: Short-range Dynamics Modeling.}
In the initial phase, we focus on stabilizing the foundational visual and short-term motion patterns. 
The model is trained on short video clips consisting of 64 frames. 
This shorter sequence length allows for larger batch sizes and faster convergence, enabling the model to efficiently learn low-level textures, spatial structures, and plausible short-term dynamics without the interference of long-range dependencies. We train the model for 200 epochs in this stage.

\noindent\textbf{Stage 2: Long-range Contextual Understanding.}
Subsequently, we extend the temporal context to capture long-term consistency and memory. 
We increase the input sequence length from 64 to 640 frames ($\sim21$ seconds of video content). 
This extension forces the Temporal Modules and Cross-Attention mechanisms to reason over extended horizons, mitigating temporal drift and maintaining spatial stability across distant frames. 
Although computationally intensive, training for 20 epochs on this extended context is sufficient to adapt the model from short-term generation to long-form video synthesis.


\subsection{Post-Training Alignment for Streaming Inference}
As illustrated in Figure~\ref{fig:framework}, during streaming inference the model conditions on clean historical frames and control signals to generate future latents. Although diffusion-forcing pre-training already equips the model with an initial ability for streaming generation, there remains a non-negligible mismatch between pre-training and practical inference. During diffusion-forcing training, each latent token is independently assigned a random noise level and the model learns to denoise under a wide range of partially corrupted historical contexts. This large training space improves the model's generative capacity and enables it to exploit ambiguous historical information. In actual streaming inference, however, we adopt a teacher-forcing-style rollout, where historical frames are clean observations with zero noise. Therefore, inference only corresponds to a restricted subset of the diffusion-forcing training distribution, which can lead to degraded generation quality and weaker temporal consistency.

To reduce this gap, we further align the model with its inference-time behavior through Distribution Matching Distillation (DMD)~\cite{yin2024one,yin2025slow}. Inspired by the second stage of self-forcing distillation~\cite{huang2026self}, we optimize the generator to directly learn how to generate future latent tokens under fixed denoising timesteps and clean historical contexts. Unlike standard self-forcing and DMD settings, our pre-trained model already supports self-rollout after diffusion-forcing pre-training. We therefore initialize the generator, the real score model, and the fake score model from the same pre-trained checkpoint. This initialization provides a stable starting point for distribution matching and avoids the instability caused by training the rollout generator from scratch.


\noindent\textbf{Randomized Score-Model Perturbation for DMD.}
A key departure from conventional DMD in our streaming pipeline is the adoption of randomized score-model perturbation, which better preserves the stochastic training behavior of Diffusion Forcing (DF) by applying independent noise to each latent token, rather than a shared global noise level. Let $\mathbf{\hat{z}}^0 = \{\mathbf{\hat{z}}_i^0\}_{i=1}^{T}$ be the denoised clean latents from the streaming rollout of the student model, and let $\mathbf{k}=\{k_i\}$, $\boldsymbol{\epsilon}=\{\boldsymbol{\epsilon}_i\}$ collect the per-frame noise steps $k_i \sim p(k)$ and Gaussian draws $\boldsymbol{\epsilon}_i \sim \mathcal{N}(\mathbf{0},\mathbf{I})$. We define the DF-consistent perturbation operator frame-wise as
\begin{equation}
    \big(\mathcal{E}_{\mathbf{k}}[\mathbf{\hat{z}}^0;\boldsymbol{\epsilon}]\big)_i = \sqrt{\bar{\alpha}_{k_i}}\,\mathbf{\hat{z}}_i^0 + \sqrt{1-\bar{\alpha}_{k_i}}\,\boldsymbol{\epsilon}_i.
\end{equation}
The perturbed sequence $\tilde{\mathbf{z}} \triangleq \mathcal{E}_{\mathbf{k}}[\mathbf{\hat{z}}^0;\boldsymbol{\epsilon}]$ is fed into the score networks together with the control condition sequence $\mathbf{c}$. We denote $\mu(\cdot)_i$ as the $i$-th frame of the score network output. With the stop-gradient operator $sg(\cdot)$, the DMD objective on latent $i$ is
{\small
\begin{equation}
    \mathcal{L}_{\text{DMD}}^i = \mathbb{E}_{\mathbf{\hat{z}}^0, \mathbf{c}, \mathbf{k}, \boldsymbol{\epsilon}} \Big[ \tfrac{1}{2}\big\| \mathbf{\hat{z}}_i^0 - sg\big( \mathbf{\hat{z}}_i^0 - \big(\mu_{real}(\tilde{\mathbf{z}}, \mathbf{c}, \mathbf{k})_i - \mu_{fake}^{\phi}(\tilde{\mathbf{z}}, \mathbf{c}, \mathbf{k})_i\big) \big) \big\|_2^2 \Big],
\end{equation}
}

where $\mu_{real}$ denotes the fixed real score network and $\mu_{fake}^{\phi}$ denotes the fake score network with parameter $\phi$. The fake score network is trained on the streaming rollout of student network with the DF loss (Eq.~\ref{eq:df}).
Empirically, this randomized perturbation yields superior generation quality over vanilla shared-noise DMD. We attribute this improvement to the alignment between the independently noised inputs and DF's pre-training distribution, which enables the teacher to provide more informative gradients for stable long-horizon streaming generation, while the decoupled noise samples for teacher and student prevent overfitting to fixed noise patterns. 

\noindent\textbf{Extended-Horizon Post-Training.}
Long-term playability requires the model to maintain stable memory over extended interactive rollouts. In post-training, we therefore further increase the training horizon as much as possible under hardware constraints. After extensive optimization, we train the model with sequences of 512 frames, corresponding to approximately 17 seconds of gameplay, on 96GB H20 GPUs. This extended horizon substantially improves the model's ability to preserve scene context, action-dependent dynamics, and temporal consistency over long periods, reducing the probability of drift or collapse during continuous play.

\begin{figure}[t]
  \centering
  \includegraphics[width=\linewidth]{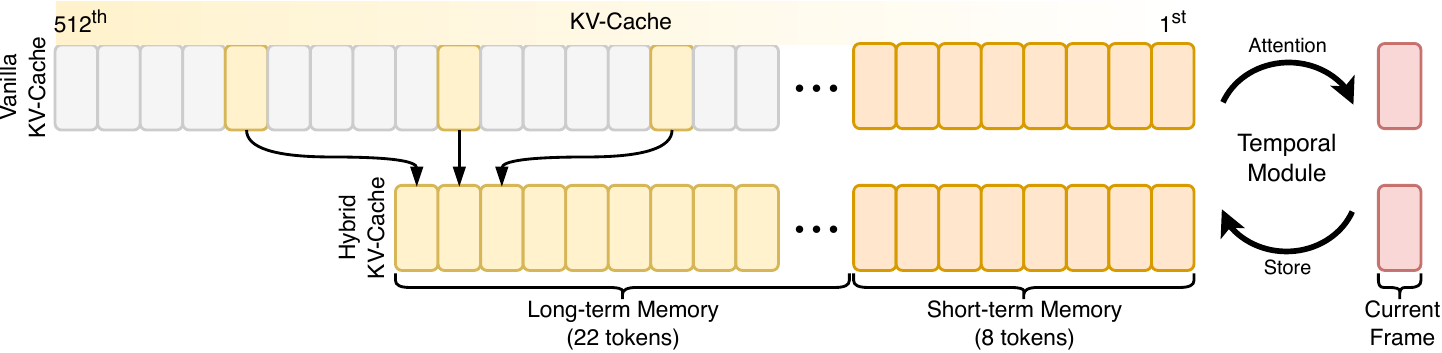}
  \caption{Overview of the Hybrid KV-Cache Memory. First row shows the Vanilla KV-cache array, second row shows the proposed Hybrid KV-Cache array.}
  \label{fig:hybrid_memory}
\end{figure}

\noindent\textbf{Hybrid KV-Cache Memory for Efficient Infinite Streaming.}
Extending the model's temporal memory also requires increasing the effective length of the KV cache during inference. However, directly retaining a dense KV cache over a long horizon significantly increases computation and memory consumption, resulting in higher latency and lower frame rate. This directly conflicts with the goal of playable low-latency generation. To address this issue, we propose a hybrid short-term and long-term KV-cache memory design that preserves long-range context while keeping the number of cached tokens small.

Specifically, we divide the KV cache into two complementary parts. The long-term memory stores sparsely sampled historical tokens, retaining global scene context and slow-changing gameplay state over an extended period. The short-term memory stores a consecutive window of the most recent tokens, preserving local temporal continuity and fine-grained motion consistency. This hybrid design enables the model to access both distant context and recent dynamics without attending to every historical token.

In our implementation, the KV cache stores 30 tokens in total. The first 22 tokens are used as long-term memory and are selected at an interval of 5 frames, while the last 8 tokens are used as short-term memory and are stored consecutively. With this design, the model can maintain an effective memory span of up to 16 seconds with a limited number of cached tokens. Importantly, we apply the same hybrid memory policy during both post-training distillation and streaming inference. This keeps the memory layout consistent between training and testing, preventing the generator from overfitting to a dense historical context that is unavailable at inference time. One visualization can be viewed at Figure~\ref{fig:hybrid_memory}.
\begin{figure}[t]
  \centering
  \includegraphics[width=0.9\linewidth]{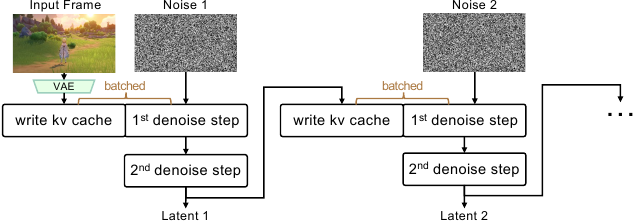}
  \caption{\textbf{Parallelized Teacher-Forcing Inference.} We merge the first denoise step and kv cache writing process to accelerate the overall inference procedure.}
  \label{fig:parallelinfer}
\end{figure}

\noindent\textbf{Parallelized Teacher-Forcing Inference.}
For streaming generation, we adopt teacher-forcing inference rather than shifted-window denoising. Since the model always conditions on finalized clean historical frames, teacher forcing reduces the ambiguity of historical context and improves generation quality. It also avoids repeatedly denoising overlapping windows, which is important for low-latency interactive gameplay. However, a naive implementation still suffers from limited parallelism because cache writing and next-frame denoising are executed sequentially.

To improve inference efficiency, we fuse the KV-cache writing step with the initial denoising step of the next frame. As shown in Figure~\ref{fig:parallelinfer}, instead of performing three separate forward passes with batch size 1, we execute one forward pass with batch size 2 followed by one forward pass with batch size 1. This batching strategy improves hardware utilization and reduces per-frame latency without changing the teacher-forcing rollout formulation. As a result, Streaming-AAA retains the quality benefits of clean-history conditioning while achieving higher inference throughput for real-time play.

\section{Experiments}
\label{sec:experiments}

\subsection{Implementation Details}

\noindent\textbf{Training.}
The VAE is optimized using a composite loss function combining $\mathcal{L}_{2}$ (MSE) for pixel-wise fidelity and LPIPS for perceptual quality. We use a learning rate of $2 \times 10^{-5}$. 
Our diffusion model is initialized from Stable Diffusion v1.5~\cite{rombach2022high}. To reduce computational overhead, we remove the two self-attention modules at the largest spatial resolution. For temporal modeling, we incorporate Motion Modules~\cite{guo2023animatediff} to perform attention along the temporal axis, where we reduce the number of temporal blocks from 2 to 1 to further improve efficiency. The resulting model has a total of 1.2B parameters.
Training proceeds in three stages. In Stage~1, we train the model with the diffusion-forcing strategy on short video clips of 64 frames ($\sim$2\,s). In Stage~2, to enhance long-term memory, we significantly extend the training sequence length to 640 frames ($\sim$21\,s). In Stage~3, we perform post-training alignment using a self-forcing-style distillation approach~\cite{huang2026self} combined with our hybrid short-long memory design, with a training length of 512 frames ($\sim$17\,s).

\noindent\textbf{Inference.}
We evaluate our model on NVIDIA A800 GPUs. To maximize throughput, the diffusion model and the VAE decoder are deployed on separate GPUs for pipelined execution. We set the number of denoising steps to 2, with specific timesteps at 999 and 666. To further reduce per-frame latency, we fuse the KV-cache writing step with the first denoising step (timestep 999) into a single batched forward pass. Additionally, we apply \texttt{torch.compile} to accelerate the diffusion model and convert the VAE decoder to TensorRT for optimized inference.

\begin{figure}[!tb]
    \centering
    \includegraphics[width=\textwidth]{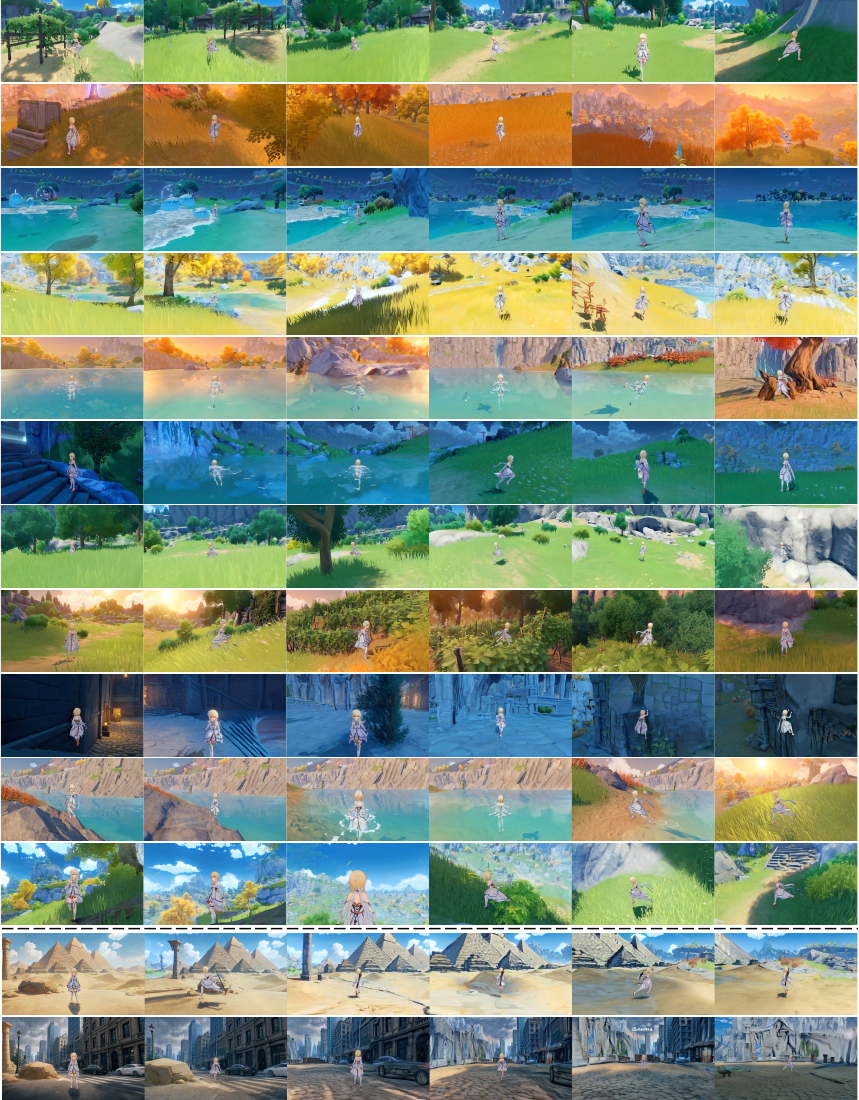}
\vspace{-1mm}
    \caption{\textbf{Qualitative results of our method on \textit{Genshin Impact}.} The top eleven rows show rollouts from in-domain frames, while the bottom two rows show rollouts from out-of-domain frames.}
\vspace{-2mm}
    \label{fig:ysresults}
\end{figure}

\subsection{Qualitative Results}

\noindent\textbf{In-domain visualization.}
The upper eleven rows of Figure~\ref{fig:ysresults} show in-domain generation results. Our method synthesizes scenes across different times of day, including daytime (first and second rows) and night (third row). It also faithfully reproduces the wild monsters that populate the game world: as shown in the third row, our model generates the cryo slimes commonly encountered in Genshin Impact. Beyond running on flat ground, the avatar further supports climbing and swimming. Together, these results demonstrate the effectiveness of our method across diverse scenarios.

\noindent\textbf{Out-of-domain visualization.}
The bottom two rows of Figure~\ref{fig:ysresults} show generation results in out-of-domain scenes, where we use FLUX~\cite{flux-2-2025} to replace the background with a pyramid and a busy city street, respectively. As illustrated in the second-to-last row, the pyramid retains its shape without noticeable distortion even as the avatar moves through the scene, and the content gradually transitions back into the in-domain distribution. This indicates that our method generalizes to a certain degree beyond the training distribution.

\begin{figure}[!tb]
    \centering
    \includegraphics[width=0.98\textwidth]{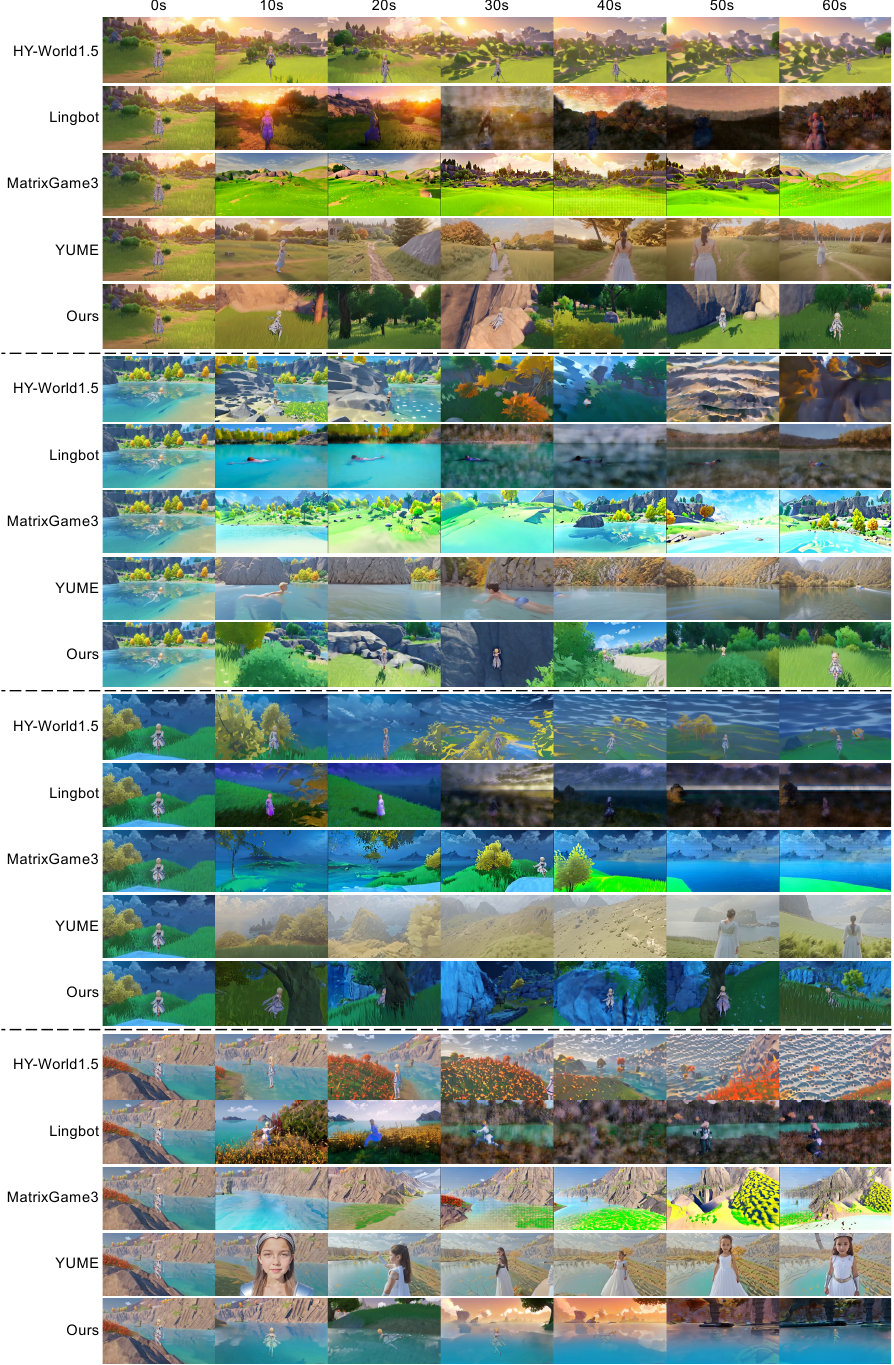}
    \vspace{-1mm}
    \caption{\textbf{Qualitative comparison with state-of-the-art approaches on \textit{Genshin Impact}.} We show our method and the competing approaches on four different cases over a one-minute rollout; rows are methods and columns are timestamps.}
    \vspace{-2mm}
    \label{fig:ys_cmp}
\end{figure}

\noindent\textbf{Comparison with state-of-the-art approaches.}
We compare our method against HY-World1.5~\cite{hyworld2025}, Lingbot~\cite{lingbot-world}, Matrix Game 3.0~\cite{2026matrix}, and YUME~\cite{mao2026yume1} on Genshin Impact. As shown in Figure~\ref{fig:ys_cmp}, our method surpasses all competing approaches in generation quality. In the first dusk scene, HY-World1.5 produces repeated shadow patterns that substantially degrade the visual result. Lingbot develops black blotchy artifacts after roughly 30\,s of rollout. Matrix Game 3.0 renders both cases with a slightly overexposed appearance. YUME drifts rapidly from the cartoon style of the source domain toward photorealism, synthesizing realistic human bodies and washing the overall palette out to gray. In contrast, our method sustains high-quality generation throughout, without quality degradation as the rollout extends, which confirms the effectiveness of our design.

\begin{figure}[t]
    \centering
    \includegraphics[width=\textwidth]{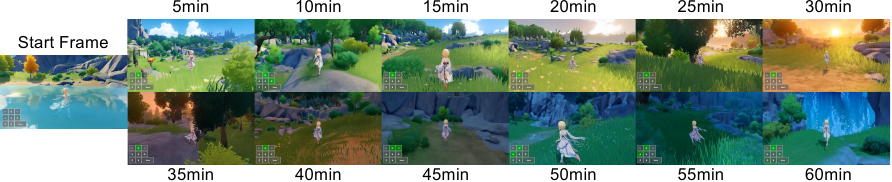}    \caption{The results of playing continuously for one hour using our model.}
    \label{fig:onehour}
\end{figure}

\noindent\textbf{Infinite-length gameplay.}
Our method is capable of supporting infinite-length gameplay. In our comparison with state-of-the-art approaches, we observe that most existing methods struggle to sustain long-horizon generation---typically, their outputs collapse within the first 30 seconds. In contrast, our method exhibits no signs of visual collapse even after one minute of continuous generation. To further verify the infinite-length gameplay capability of our method, we conduct a stress test by playing continuously for one hour. As shown in Figure~\ref{fig:onehour}, throughout the entire hour of gameplay, our method neither encounters catastrophic collapse that halts generation nor suffers from gradual visual degradation as the play duration extends. These results clearly demonstrate the infinite-length gameplay capability of our approach.

\begin{table}[!tb]
\centering
\caption{\textbf{Quantitative and efficiency comparison with state-of-the-art methods.} We compare our method against existing state-of-the-art approaches on several VBench metrics, together with output resolution, frame rate, and action-to-pixel latency. Best result per column is in bold.}
\label{tab:quantitative_ys}
\resizebox{\linewidth}{!}{
\begin{tabular}{lccccc|ccc}
\toprule
& \multicolumn{5}{c|}{Generation Quality} & \multicolumn{3}{c}{Efficiency} \\
\cmidrule(lr){2-6} \cmidrule(lr){7-9}
Method & Subject Consist. & Background Consist. & Motion Smooth. & Imaging Quality & Average & Resolution & Frame Rate ($\uparrow$) & Latency ($\downarrow$) \\
\midrule
Lingbot      & 0.689  & 0.876  & 0.983  & 0.438  & 0.747  & 720p & 2.6 FPS~(8GPUs) & 8200 ms \\
YUME          & 0.716  & 0.860  & 0.985  & 0.639  & 0.800  & 720p & 6.7 FPS & 6600 ms \\
Matrix Game 3.0 & 0.690  & 0.887  & 0.987  & 0.666  & 0.807  & 720p & 6.1 FPS & 9800 ms \\
HY-World1.5    & 0.755 & 0.906 & \textbf{0.988} & 0.589 & 0.810 & 480p & 2.1 FPS & 11300 ms \\
Ours          & \textbf{0.761} & \textbf{0.910}  & 0.982  & \textbf{0.677} & \textbf{0.832} & \textbf{1080p} & \textbf{43.0 FPS} & \textbf{198 ms} \\
\bottomrule
\end{tabular}
}
\end{table}

\subsection{Quantitative Results}

Following the protocol of Section~\ref{sec:experiments}, we evaluate all methods with VBench~\cite{huang2023vbench,huang2025vbench++,zheng2025vbench2} in custom\_input mode on subject consistency, background consistency, motion smoothness, and imaging quality, and report in Table~\ref{tab:quantitative_ys} the scores averaged over both evaluation titles.

As shown in Table~\ref{tab:quantitative_ys}, our method outperforms all competing approaches---HY-World1.5~\cite{hyworld2025}, Lingbot~\cite{lingbot-world}, Matrix Game 3.0~\cite{2026matrix}, and YUME~\cite{mao2026yume1}---on subject consistency, background consistency, and imaging quality, and leads by a clear margin on the average score. Compared with HY-World1.5, we are on par on subject consistency and background consistency, but substantially better on imaging quality (0.677 vs. 0.589). Figure~\ref{fig:ys_cmp} explains the gap: beyond roughly 30\,s of rollout HY-World1.5 develops pronounced repetitive patterns, which severely hurt imaging quality. Our motion smoothness is marginally lower than that of the baselines. A plausible explanation is that the baselines move very slowly, so their rollouts consist of comparatively gentle frame-to-frame changes, whereas our rollouts contain fast locomotion and abrupt events such as jumping off a cliff, which naturally depress this metric. Our advantage is even more pronounced in efficiency: we run about $20\times$ faster than HY-World1.5 while cutting latency by more than $98\%$, and we do so at 1080p rather than 480p--720p. Taken together, these results show that our method dominates existing state-of-the-art approaches in both generation quality and inference efficiency.

\begin{figure}[!tb]
    \centering
    \includegraphics[width=\textwidth]{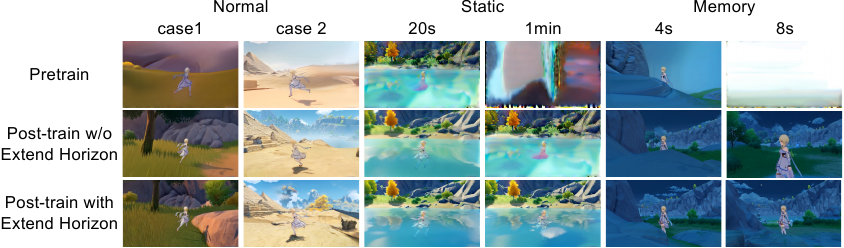}
    \caption{\textbf{Ablation study on \textit{Genshin Impact}.} We compare the pre-trained model against post-training with a short and an extended training horizon across three scenarios: \textit{Normal} navigation from in-domain (Case~1) and out-of-domain (Case~2) starting frames; \textit{Static} standstill for 20\,s and 60\,s; and \textit{Memory} retention after turning the camera away for 4\,s/8\,s and back.}
    \vspace{-2mm}
    \label{fig:ys_ablation}
\end{figure}

\subsection{Ablation Study}

Figure~\ref{fig:ys_ablation} presents our ablation study, comparing the pre-trained model against post-training with a short and an extended training horizon.

\noindent\textbf{Normal scenes.}
Case~1 and Case~2 report results from in-domain and out-of-domain starting frames, respectively. The pre-trained model degenerates within the first 5\,s of rollout, with the generated content collapsing toward a monotonous appearance, whereas both post-trained models stay stable in either case. Post-training therefore yields a substantial quality gain within our framework.

\noindent\textbf{Static scenes.}
When the avatar is not moved, the pre-trained model already breaks down after 20\,s, and post-training with a short horizon becomes blurry after 60\,s. Only the extended-horizon variant preserves a sharp and stable scene throughout.

\noindent\textbf{Memory.}
We rotate the camera away and then turn back to the original viewpoint. With a short post-training horizon, the scene is already slightly altered after 4\,s and differs substantially after 8\,s; with the extended training horizon and longer training, it remains essentially unchanged even after 8\,s. These comparisons show that both extending the post-training horizon and increasing training length are necessary to confer markedly stronger resistance to collapse in static scenes and a genuine degree of memory.

\subsection{Efficiency and Latency Analysis}

Frame rate and action-to-pixel latency are critical to interactive experience. End-to-end latency comprises the diffusion model (93\,ms on an A800) and the VAE decoder (58\,ms). Since player actions may arrive mid-execution, we budget an expected extra half-step of diffusion, giving $93 \times 1.5 + 58 \approx 198$\,ms; this is the theoretical computation latency, excluding network transmission and codec overhead. Deploying the diffusion model and VAE decoder on separate GPUs for pipelined execution further yields $1000 / 93 \times 4 \approx 43$\,fps, enabling real-time interactive gameplay. Relative to existing streaming systems, which operate at 480p--720p with single-digit frame rates and multi-second latency, our system delivers 1080p output at 43\,fps with 198\,ms latency; the detailed comparison is reported in Table~\ref{tab:quantitative_ys}.

\section{Conclusion}

We have presented \textbf{Streaming-AAA}, a controllable infinite AAA game video generation framework that achieves high-resolution (1080p), high-frame-rate ($\sim$43\,fps), and low-latency ($\sim$200\,ms) streaming gameplay synthesis. Our framework integrates an efficient high-resolution video VAE with pixel-shuffle decoding, a factorized UNet-based latent diffusion model with unified multi-modal conditioning, and a progressive training pipeline spanning short-range dynamics modeling, long-context pre-training, and post-training alignment via distribution matching distillation. The proposed hybrid short-term and long-term KV-cache memory, combined with parallelized teacher-forcing inference, enables stable infinite-horizon generation while maintaining real-time interactivity. Experiments on \textit{Genshin Impact} demonstrate that Streaming-AAA produces visually rich and temporally consistent gameplay across diverse environments, weather conditions, and character actions.

\noindent\textbf{Limitations.}
Our current approach is primarily designed and optimized for playability within a specific game title. The model is trained on domain-specific data from a single game, and its generalization to unseen game environments, art styles, or game mechanics remains limited. Extending the framework to a universal game generation model that handles diverse titles without per-game fine-tuning is an important open challenge.

\noindent\textbf{Future Work.}
Looking ahead, we believe the core techniques developed in Streaming-AAA---efficient high-resolution VAE, controllable streaming diffusion, and long-horizon memory---can serve as building blocks for a broader range of downstream applications beyond playable generation. For instance, the framework could be adapted for \textit{game-specific visual enhancement}, where the model upscales or refines low-quality rendered frames into high-fidelity outputs in real time. We hope that Streaming-AAA provides a solid foundation toward bridging generative video models and practical interactive entertainment systems.

{
    \small
    \bibliographystyle{ieeenat_fullname}
    \bibliography{main}
}


\end{document}